\documentclass{article} %
\usepackage{iclr2019_conference,times}
\usepackage[utf8]{inputenc}
\usepackage[T1]{fontenc}

\usepackage{amsmath,amsfonts,bm}

\def\eqref#1{equation~\ref{#1}}

\def\1{\bm{1}}

\DeclareMathAlphabet{\mathsfit}{\encodingdefault}{\sfdefault}{m}{sl}
\SetMathAlphabet{\mathsfit}{bold}{\encodingdefault}{\sfdefault}{bx}{n}

\usepackage[hidelinks]{hyperref}
\usepackage{url}
\usepackage[inline]{enumitem}
\usepackage{xspace}
\usepackage{graphicx}
\usepackage{adjustbox}

\graphicspath{ {images/} }

\newcommand{\ie}{\textit{i.e.,}\xspace}
\newcommand{\eg}{\textit{e.g.,}\xspace}
\newcommand{\etc}{\textit{etc.}\xspace}

\title{A Literature Study of Embeddings on Source Code}

\author{Zimin Chen \& Martin Monperrus \\
School of Electrical Engineering and Computer Science\\
KTH Royal Institute of Technology\\
\texttt{zimin@kth.se, martin.monperrus@csc.kth.se}
}

\iclrfinalcopy %

\begin{document}

\maketitle

\begin{abstract}
Natural language processing has improved tremendously after the success of word embedding techniques such as word2vec. Recently, the same idea has been applied on source code with encouraging results. 
In this survey, we aim to collect and discuss the usage of word embedding techniques on programs and source code.
The articles in this survey have been collected by asking authors of related work and with an extensive search on Google Scholar.
Each article is categorized into five categories:
\begin{enumerate*}
\item embedding of tokens
\item embedding of functions or methods
\item embedding of sequences or sets of method calls
\item embedding of binary code
\item other embeddings.
\end{enumerate*}
We also provide links to experimental data and show some remarkable visualization of code embeddings.
In summary, word embedding has been successfully applied on different granularities of source code. 
With access to countless open-source repositories, we see a great potential of applying other data-driven natural language processing techniques on source code in the future.
\end{abstract}

\section{Introduction}

With recent success in deep learning, it has become more and more popular to apply it on images, natural language, audios and \etc 
One key aspect of this research is to compute a numerical vector representing the object of interest: this vector is called an ``embedding''. For words in natural language, researchers have come up with a set of techniques, called word embedding, that maps words or phrases to a vector of real numbers. 

Moving from natural language words and sentences to programming language tokens and statements is natural. It is also meaningful because \cite{hindle2012naturalness} showed that programming languages are like natural languages, with comparable repetitiveness and predictability . In this line of thought, authors have envisioned potentially powerful applications of word embeddings on source code, and experimental results are encouraging. This paper presents this new research area on embeddings on source code.

Much like how natural languages have characters, words, sentences and paragraphs, programming languages also have different granularities, such variables, expressions, statements and methods. Therefore, word embeddings can be applied to source code on different granularities.

In this paper, we collect and discuss the usage of word embeddings in programs. Our methodology is to identify articles that compute and use an embedding on source code. We do that with asking authors of related work directly and extensive search with Google Scholar.

Our contributions are:
\begin{itemize}
\item A survey of word embeddings on source code in five categories: embeddings of tokens, embeddings of expressions, embeddings of APIs, embeddings of methods, and other  miscellaneous embeddings.
\item A collection of visual representations of embeddings, that add an aesthetic dimension to the power of code embeddings.
\item A curated list of publicly available code embedding data.
\end{itemize}
\section{Background}

\subsection{Embeddings}
An embedding is mapping from objects to vectors of real numbers.
Word embedding refers to all natural language processing approaches where words (or sequence of words) are mapped onto vectors of real numbers. It makes it possible to work with textual data in a mathematical model. It also has the advantage that fundamentally discrete data (words) is transformed into continuous vectors space. With recent progress in word embedding, especially word2vec (\cite{mikolov2013distributed}), the embedding vectors preserve the semantic. Word embedding such as word2vec requires large unlabeled corpora to train on.

\subsection{Visualization}
Embedding vectors usually have more than 3 dimensions, up to hundreds and even thousands of dimensions. This means we have to reduce the dimensionality in order to visualize them in a 2D plane or 3D space. The most commonly used techniques are Principal Component Analysis (PCA, \cite{pearson1901liii}) and t-Distributed Stochastic Neighbour Embedding (t-SNE, \cite{maaten2008visualizing}).

\subsection{Similarity Metrics in the Embedding Space}
Similarity between words can be measured by calculating a similarity metric between their embeddings. The most used similarity metric for word embeddings is cosine similarity. The cosine similarity measures the angle between two vectors, which is independent of their magnitude. Other similarity metrics exists such as Euclidean distance, but in high dimensional space where the data points are sparse, \cite{beyer1999nearest} showed that the ratio of distance between the nearest and farthest point is close to 1. And \cite{aggarwal2001surprising} proved that between $L_{k}$ norms, smaller k is more preferable in high dimensional space, \ie $L_{1}$ norm (Manhattan distance) is more preferable than $L_{2}$ norm (Euclidean distance).

\section{Embedding for Source code}
Here we present source code embeddings, each subsection focusing a specific granularity.
Those different granularities are needed depending on the downstream task, \eg we need token embedding in code completion and we need function embedding in function clone detection. \autoref{tab:link} shows links to publicly available experimental data.

\subsection{Embedding of Tokens}

\cite{harer2018automated} use word2vec to generate word embedding for C/C++ tokens for software vulnerability prediction. The token embedding is used to initialize a TextCNN model for classification.

\cite{white2017sorting} generate Java token embedding using word2vec for automatic program repair. They used the embedding to initialize a recursive encoder of abstract syntax trees.

\cite{azcona2019user2code2vec} explore Python token embeddings for profiling student submissions. Each student's submission is transformed into a flattened list of token embeddings, and then each student is represented as one matrix where each row represents one submission.

\cite{chen2018remarkable} use word2vec to generate java token embedding for finding the correct ingredient in automated program repair . They use cosine similarity on the embeddings to compute a distance between pieces of code.

\subsection{Embedding of Functions or Methods}

\begin{figure}
    \centering
    \includegraphics[width=\linewidth]{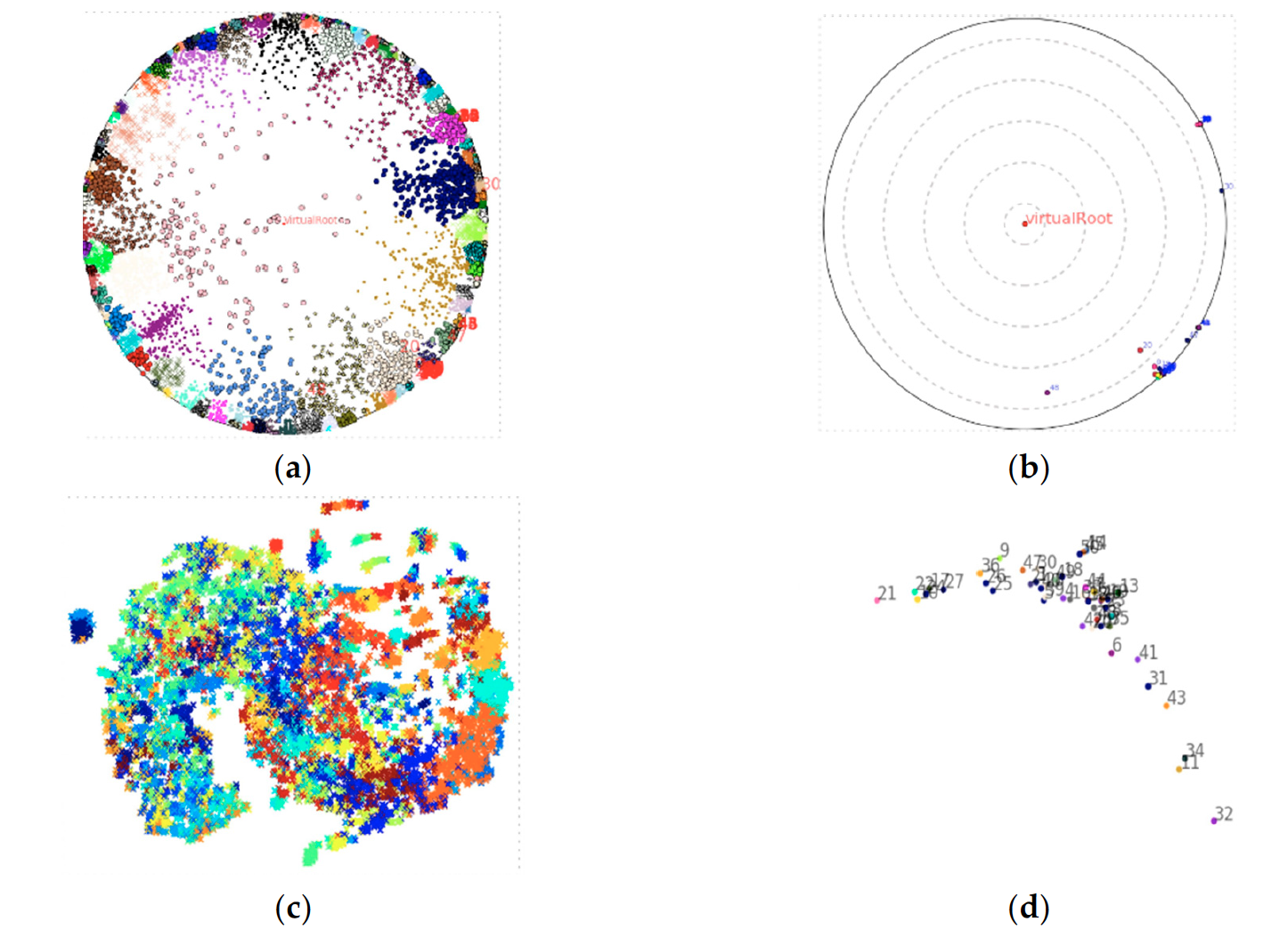}
    \caption{Function embedding from \cite{lu2019hyperbolic}, reproduced with permission.}
    \label{fig:Lu_1}
\end{figure}

\cite{alon2019code2vec} compute Java method embeddings for predicting method names. Abstract syntax tree is used to construct a path representation between two leaf nodes. Bag of path representations are all aggregated into one single embedding for a method. In a subsequent paper \cite{alon2018code2seq}, the path representation is computed using Long Short Term Memory network (LSTM) instead of a single layer neural network.

\cite{allamanis2015suggesting} use a logbilinear context model to generate an embedding for method names. They define a local context which captures tokens around the current token, and a global context which is composed of features from the method code. Their technique is able to generate new method names never seen in the training data.

\cite{defreez2018path} generate function embeddings for C code using control-flow graphs. They perform a random walk on interprocedural paths in the program, and used the paths to generate function embeddings. The embedding is used for detecting function clones.

\cite{murali2017neural} compute a function embedding for program sketches. They use the embedding to generate source code given a specification that lists API calls and API types to be used in the generated code. An encoder-decoder model is used to generate program sketches from the specification. The final hidden state of the encoder is seen as embedding for the function.

\cite{lu2019hyperbolic} propose a new function embedding method that learns embeddings in a hyperbolic space. They first construct function call graphs where the weight of edges are replaced with a Ricci curvature. The embeddings are then learned by using a Poincaré model. The function embedding in is visualized in \autoref{fig:Lu_1}.

\cite{devlin2017semantic} use function embedding as input to repair variable misuse in Python. They encode the function AST by doing a depth first traversal and create an embedding by concatenating the absolute position of the node, the type of node, the relationship between the node and its parent and the string label of the node.

\cite{8668039} generate embedding for methods to do code clone detection, by jointly considering embeddings of AST node type and of node contents. The AST to be embedded is traversed from leaf nodes to the root, and the obtained sequences are fed to a LSTM for learning sequences. The similarity metric between methods is computed using a Siamese network.

\subsection{Embedding of Sequences or Sets of Method Calls}

\begin{figure}
    \centering
    \includegraphics[width=\linewidth]{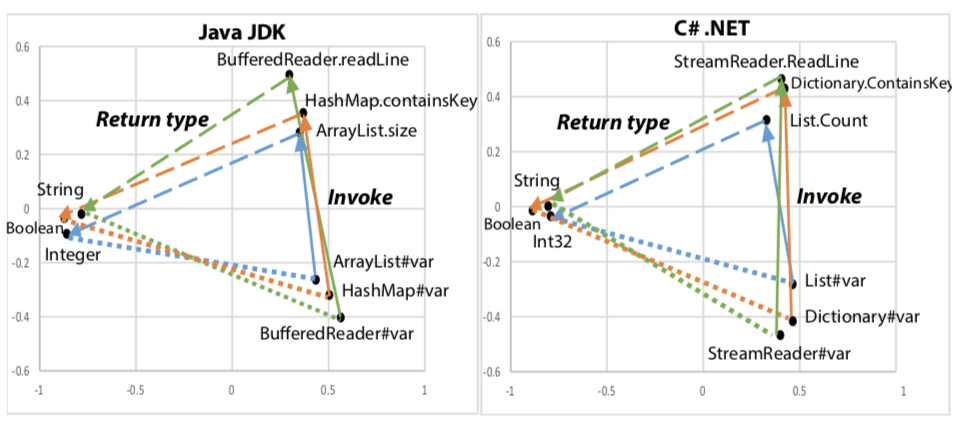}
    \caption{API embedding from \cite{nguyen2016mapping}, reproduced with permission.}
    \label{fig:Nguyen_2}
\end{figure}

\cite{henkel2018code} generate embedding for abstracted symbolic traces from C projects. The traces are collected from symbolic executions, and then names in symbolic traces are renamed to reduce the vocabulary size. The embedding is learned from the abstracted symbolic traces using word2vec.

\cite{nguyen2016mapping} explore embeddings for Java and C\# APIs, and used it to find similar API usage between the languages. They use word2vec for generating API element embeddings, and based on known API mappings, they can find cluster with similar usage across two languages. The API embedding is visualized in \autoref{fig:Nguyen_2}. In a subsequent paper \cite{nguyen2017exploring}, they translate source code to API sequences, and learned API embedding from abstracted API sequences using word2vec.

\cite{gu2016deep} consider the problem of translating natural language query to a sequence of API calls. They use an encoder-decoder model with custom a loss function, where the input is the natural language query and output is the sequence of API calls. The final hidden state of the encoder can be seen as an embedding for the natural language query.

\cite{pradel2018deepbugs} propose to use code snippet embeddings to identify potential buggy code. To generate the training data, they collect code examples from a code corpus and apply simple code transformation to insert an artificial bug. Embedding for positive and negative code examples are generated and they are used to train a classifier.

\subsection{Embedding of Binary Code}

\begin{figure}
    \centering
    \includegraphics[width=\linewidth]{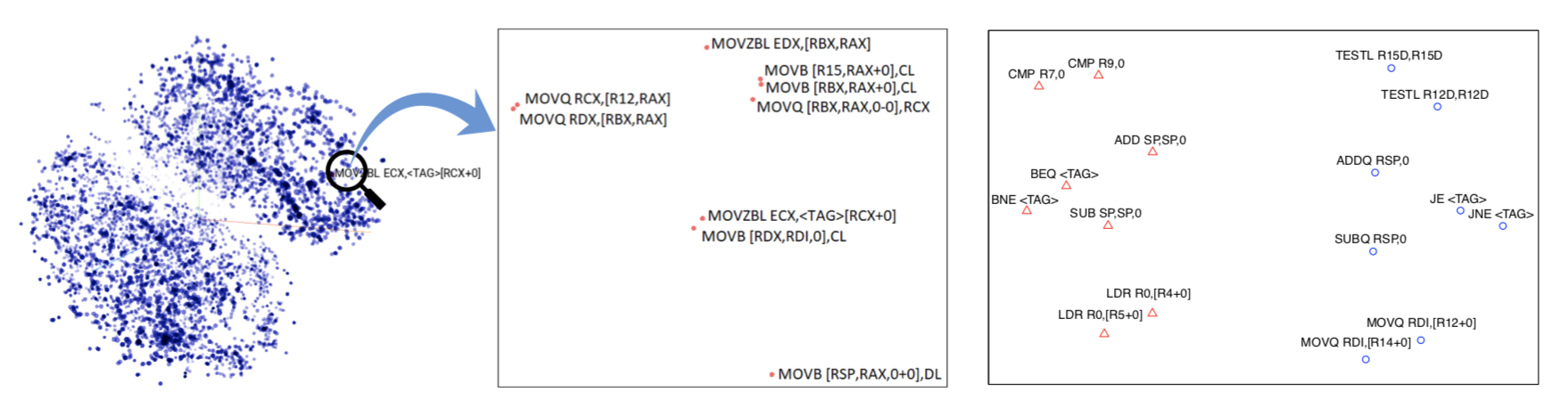}
    \caption{Binary instruction embedding from \cite{zuo2018neural}, reproduced with permission.}
    \label{fig:Zuo_1}
\end{figure}

\cite{ben2018neural} use the LLVM Intermediate Representation (IR) to build a contextual flow graph, which is used to generate binary code embedding for C/C++ statements. The source code is first compiled to LLVM IR, where the identifier and literals are renamed. The abstracted representation is used to build a contextual flow graph, where paths are extracted for training the embeddings.

\cite{zuo2018neural} compute binary instruction embedding to calculate the similarity between two binary blocks. Binary block embeddings are trained on x86-64 and ARM instructions, and instructions in a basic block are combined using LSTM. A Siamese architecture is adopted to determine the similarity between two basic blocks. The binary instruction embedding is visualized in \autoref{fig:Zuo_1}.

\cite{redmond2018cross} explore binary instruction embedding across architectures. Binary instruction embeddings are learned by cluster similar instructions in the same architecture, and preserve the semantic relationship between binary instruction in different architectures.

\cite{xu2017neural} generate cross-platform binary function embedding for similarity detection. The binary function is converted to control flow graph and embedding is generated using Structure2vec. Then, Siamese network is trained for similarity detection.

\subsection{Other Embeddings}

\begin{figure}
    \centering
    \includegraphics[width=\linewidth]{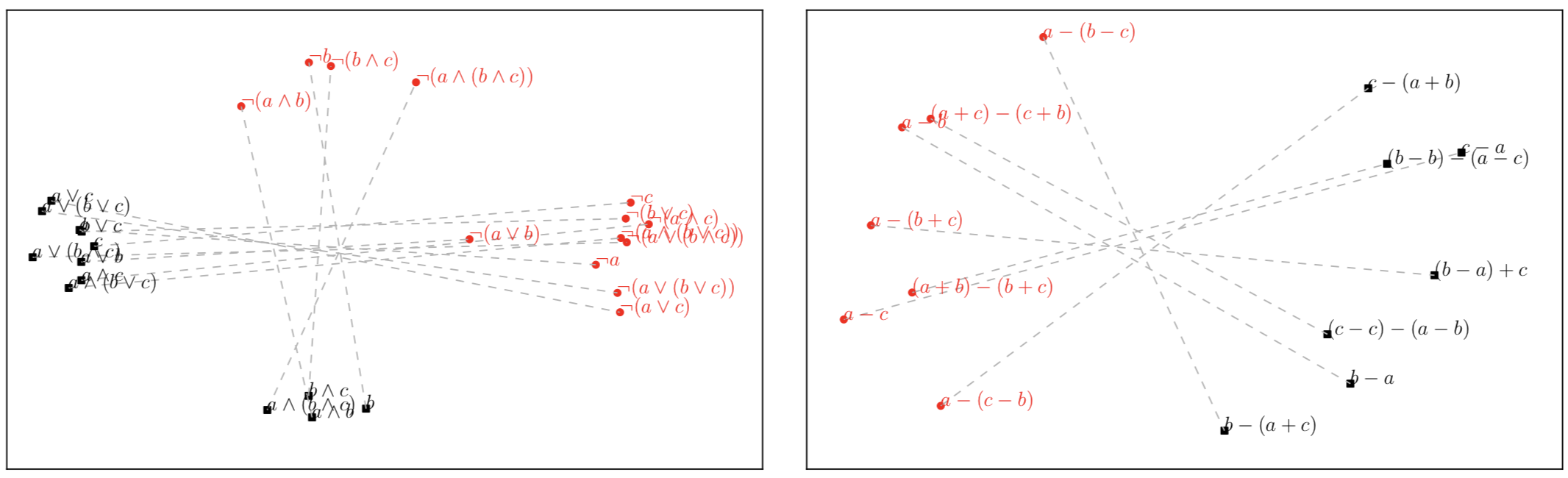}
    \caption{Expression embedding from \cite{allamanis2017learning}, reproduced with permission.}
    \label{fig:Allamanis_1}
\end{figure}

\cite{allamanis2017learning} use TreeNN to generate embedding for algebraic and logical expressions. Their technique recursively encodes non-leafs node in the abstract syntax tree structure by combining the children nodes using a multi-layer perceptron. The expression embedding is visualized in \autoref{fig:Allamanis_1}.

\cite{yin2018learning} introduce the problem of learning embedding for edits over source code, \ie patches. The edits are represented as sequence or graph, and a neural network is used to generate the embedding.

\cite{zhao2018neural} use an embedding approach to determine if two Android components have an Inter-Component Communication link. An intent is a message sent between different Android components, and an intent filter describes what types of intent the component is willing to receive. The intent and filter embeddings are used to train a classifier for possible Inter-Component Communication link. 

\cite{wang2017dynamic} generate program embedding using execution traces. The program is represented as multiple variable traces, and they are interleaved to simulate dependencies between traces. A full program embedding is the average over all embeddings of the traces.

\cite{chistyakov2018semantic} extract behaviour patterns from a sequence of system events, and combine them to an embedding that represents the program behaviour. First, they generate an embedding using an autoencoder, and then all behaviour patterns are combined using min, max and mean functions over elements.

\cite{allamanis2017graph} use a graph neural network to generate embedding for each node in an abstract syntax tree. The graph is constructed from nodes and edges from the AST, and then enriched with additional edges indicating data flow and type hierarchies.

\cite{Theeten2019import2vec} explore software library embedding for Python, Java and JavaScript. They collected library usage in open source repositories and ignored the import order. The library embedding is trained by consider all imported libraries in the same file as context. 

\begin{table}
\begin{center}
\begin{adjustbox}{width=\textwidth}
\begin{tabular}{ll}
\hline
\cite{white2017sorting}   & \url{https://sites.google.com/view/deeprepair/home} \\
\cite{azcona2019user2code2vec} & \url{https://github.com/dazcona/user2code2vec} \\
\cite{chen2018remarkable} & \url{https://github.com/kth-tcs/3sFix-experiments} \\
\cite{alon2019code2vec} & \url{https://github.com/tech-srl/code2vec} \\
\cite{alon2018code2seq} & \url{http://github.com/tech-srl/code2seq} \\
\cite{allamanis2015suggesting} & \url{http://groups.inf.ed.ac.uk/cup/naturalize/} \\
\cite{defreez2018path} & \url{https://github.com/defreez-ucd/func2vec-fse2018-artifact} \\
\cite{murali2017neural} & \url{https://github.com/capergroup/bayou} \\
\cite{devlin2017semantic} & \url{https://iclr2018anon.github.io/semantic_code_repair/index.html.} \\
\cite{henkel2018code} & \url{https://github.com/jjhenkel/code-vectors-artifact} \\
\cite{nguyen2017exploring} & \url{http://home.eng.iastate.edu/~trong/projects/jv2cs/} \\
\cite{gu2016deep} & \url{https://guxd.github.io/deepapi/} \\
\cite{pradel2018deepbugs} & \url{https://github.com/michaelpradel/DeepBugs} \\
\cite{zuo2018neural} & \url{https://nmt4binaries.github.io/} \\
\cite{ben2018neural} & \url{https://github.com/spcl/ncc} \\
\cite{redmond2018cross} & \url{https://github.com/nlp-code-analysis/cross-arch-instr-model} \\
\cite{xu2017neural} & \url{https://github.com/xiaojunxu/dnn-binary-code-similarity} \\
\cite{allamanis2017learning} & \url{http://groups.inf.ed.ac.uk/cup/semvec/} \\
\cite{yin2018learning} & \url{https://github.com/Microsoft/msrc-dpu-learning-to-represent-edits} \\
\cite{wang2017dynamic} & \url{ https://github.com/keowang/dynamic-program-embedding} \\
\cite{Theeten2019import2vec} & \url{https://zenodo.org/record/2546488#.XKHb2S97GL4} \\
\hline
\end{tabular}
\end{adjustbox}
\caption{Publicly available embeddings on code.}
\label{tab:link}
\end{center}
\end{table}

\section{Future Research Directions}
Now, we speculate about potentially interesting future research directions on embedding for code.

\subsection{Embedding versus Downstream Tasks}
Code embeddings should capture the semantics, but it is a difficult task to assess its quality in a general manner. One way is to use analogies ( $a$ is to $b$, what $x$ is to $y$ ). Another way could be to list top-n closest elements in an embedding space. 
Yet, the state-of-the-art in NLP suggests to evaluate code embedding techniques on downstream tasks rather than in a generic way. The existing embeddings can be used as initialization for the downstream tasks.

\subsection{Contextual Word Embedding}
Word embedding techniques have the drawback of having a one-to-one mapping between elements and their embeddings. Consider the Java keyword 'static', which have different meaning depending on put on a field or a method: it should also have two different embeddings for those two different contexts. Contextual word embedding is a technique that takes the context around the word into account, approach like BERT by \cite{devlin2018bert} has achieved state-of-the-art results in several NLP tasks.
Considering how powerful contextual word embedding is, and how the meaning of code elements can be contextual, we believe that it is a promising direction machine learning on code. We note that \cite{allamanis2017graph} can be seen as an implicit contextual word embedding, even if it does not use this terminology. 

\section{Conclusion}
We have collected and discussed articles that define or use embedding on source code.
This literature study shows how diverse embedding techniques are, and they are applied to different downstream tasks.
We think that the field of embedding on code is only at the beginning and we expect that many papers will use this key and fascinating concept in the future.\\\
The authors would like to thank Jordan Henkel, Miltos Allamanis and Hugo Mougard for valuable pointers and discussions.

\bibliography{iclr2019_conference}
\bibliographystyle{iclr2019_conference}

\end{document}